\documentclass[runningheads]{llncs}

\usepackage{eccv}

\usepackage{eccvabbrv}

\usepackage{graphicx}
\usepackage{booktabs}

\usepackage[accsupp]{axessibility}  

\usepackage[pagebackref,breaklinks,colorlinks]{hyperref}

\usepackage{orcidlink}
\usepackage{amsmath}
\usepackage{amsfonts}
\usepackage[ruled]{algorithm2e}
\usepackage{tcolorbox}

\begin{document}

\title{Global-Local Collaborative Inference with LLM for Lidar-Based Open-Vocabulary Detection} 

\titlerunning{Global-Local Collaborative Inference}

\author{Xingyu Peng\inst{1} \and
Yan Bai\inst{2} \and Chen Gao\inst{4} \and Lirong Yang\inst{2} \and Fei Xia\inst{2} \and Beipeng Mu\inst{2} \and Xiaofei Wang\inst{2} \and Si Liu\inst{1,3}\thanks{Corresponding author}}

\authorrunning{X.~Peng et al.}

\institute{Institute of Artificial Intelligence, Beihang University \and
Meituan \and
Hangzhou Innovation Institute, Beihang University \and 
University of Science and Technology Beijing
\\
\email{\{pengxyai,liusi\}@buaa.edu.cn} \\
\email{\{baiyan02,yanglirong,wangxiaofei19\}@meituan.com} \\
\email{\{gaochen.ai,xia.fei09,mubeipeng\}@gmail.com}}
\maketitle

\begin{abstract}
  
Open-Vocabulary Detection (OVD) is the task of detecting all interesting objects in a given scene without predefined object classes. Extensive work has been done to deal with the OVD for 2D RGB images, but the exploration of 3D OVD is still limited. Intuitively, lidar point clouds provide 3D information, both object level and scene level, to generate trustful detection results. However, previous lidar-based OVD methods only focus on the usage of object-level features, ignoring the essence of scene-level information. In this paper, we propose a Global-Local Collaborative Scheme (GLIS) for the lidar-based OVD task, which contains a local branch to generate object-level detection result and a global branch to obtain scene-level global feature. With the global-local information, a Large Language Model (LLM) is applied for chain-of-thought inference, and the detection result can be refined accordingly. We further propose Reflected Pseudo Labels Generation (RPLG) to generate high-quality pseudo labels for supervision and Background-Aware Object Localization (BAOL) to select precise object proposals. Extensive experiments on ScanNetV2 and SUN RGB-D demonstrate the superiority of our methods. Code is released at \url{https://github.com/GradiusTwinbee/GLIS}.
\end{abstract}

\section{Introduction}
\label{sec:intro}

\begin{figure}[tb]
  \centering
  \includegraphics[height=6.8cm]{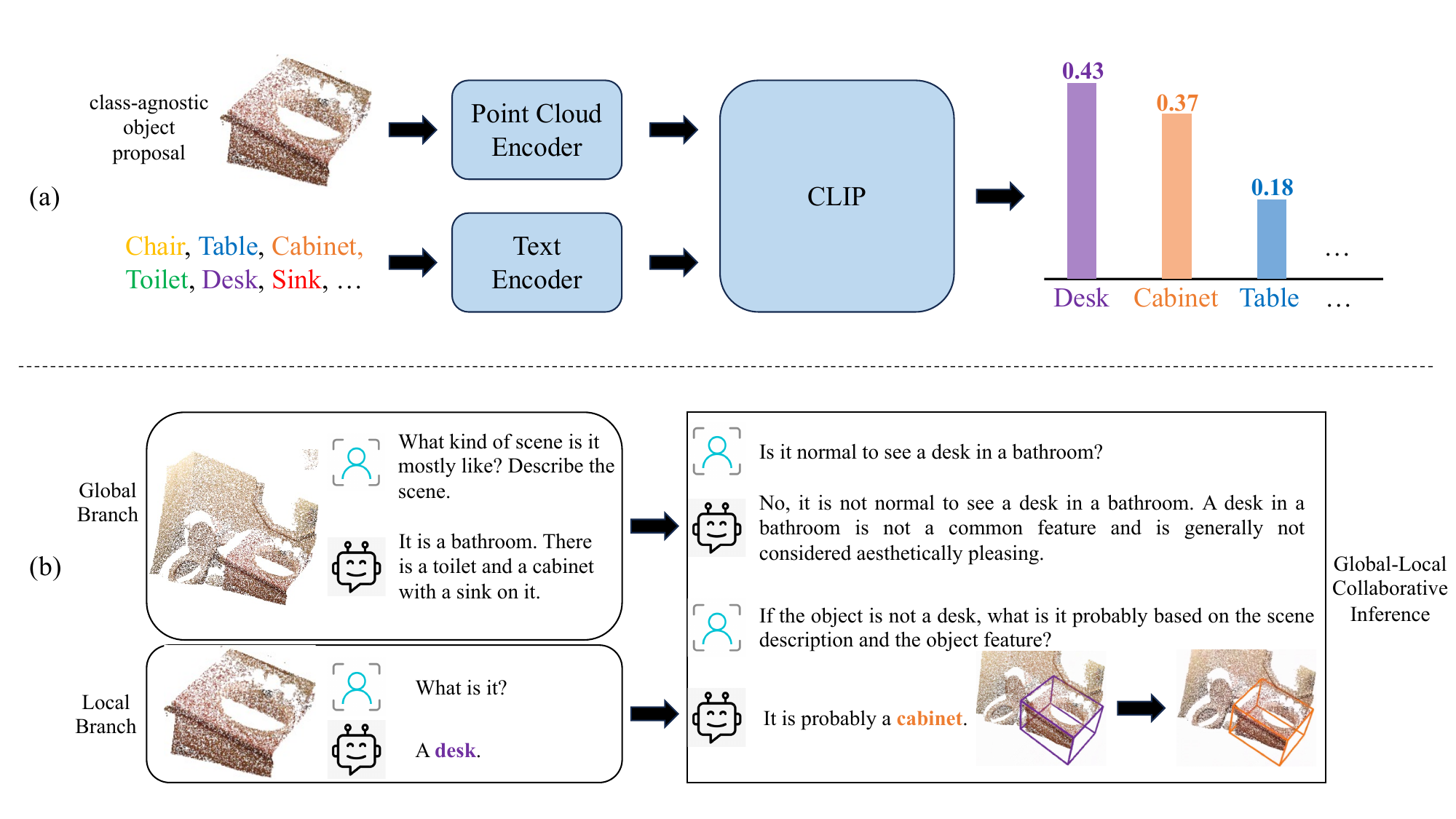}
  \caption{
  (a) The previous 3D OVD paradigm determines the class of an object proposal by comparing its point cloud feature with the class text features. As it only considers object-level/local information, the proposal may be wrongly recognized, \eg mistaking the cabinet for a desk. 
  (b) In contrast, we propose a Global-Local Collaborative Inference Scheme (GLIS) for 3D OVD, considering both the scene-level/global information and the object-level/local information. Additionally, we leverage the LLM to provide common sense for chain-of-thought inference, which can clarify the ambiguous object class step by step.
  }
  \label{fig:intro}
\end{figure} 

As a basic function of machine perception, object detection has attracted much attention within computer vision communities. The traditional training pipeline for the detection model relies on elaborately labeled data, resulting in a limited number of classes that can be collected and annotated.
In this way, the detection model fails to detect objects not belonging to the training object classes. In recent years, open-vocabulary detection has become a popular topic, which is hopeful to solve the problem. Generally, an open-vocabulary detection model requires no human-labeled data in the training stage and possesses the ability to detect any interested object in a given scene. However, although many open-vocabulary detection methods focus on 2D RGB images, the 3D point clouds, a data modality widely used in autonomous driving and robot vision, have not been thoroughly investigated. In this paper, we focus on the task of lidar-based open-vocabulary object detection, where only point clouds are utilized in the testing stage.

Compared to open-vocabulary detection for 2D RGB images, lidar-based open-vocabulary detection suffers from more difficulties. Firstly, point clouds have lower resolutions compared to 2D RGB images, leading to the loss of object details, \eg material, texture, color, \etc. Additionally, the quality of the point cloud could be easily affected by environmental conditions, thus noisy points may exist in the point cloud. Therefore, it is harder for a lidar-based detection model to recognize the class only based on the lidar information of a single object, emphasizing the necessity of environmental information in lidar-based OVD.

However, the dominant paradigm of current state-of-the-art lidar-based OVD methods only focuses on object-level features and neglects the importance of scene-level information. As shown in \cref{fig:intro}\textcolor{red}{a}, the current paradigm determines the class of an object by comparing the object-level features and the text features of class names. Such a paradigm could be easily affected by the point cloud's low resolution and noises since it only considers local information. 

To address this issue, we propose a Global-Local Collaborative Inference Scheme (GLIS) for the lidar-based OVD task, considering both the global scene-level information and the local object-level information. Additionally, the Large Language Model (LLM) is introduced to provide common sense for chain-of-thought inference. The common sense from LLM could help the model clarify the ambiguous object features, while chain-of-thought inference allows the model to refine the detection result step by step. 
For example, in the case of \cref{fig:intro}, the class-agnostic proposal, which is a cabinet in fact, is wrongly recognized as a desk. The previous paradigm (shown in \cref{fig:intro}\textcolor{red}{a}) does not perceive this error as it only considers local information. Contrastly, in GLIS (shown in \cref{fig:intro}\textcolor{red}{b}), such error could be corrected with the clues from global information. Specifically, with the scene being identified as a bathroom, the LLM recognizes that a desk is unlikely to be present, suggesting the object proposal might have been misclassified. Drawing on both local and global features, as well as insights from previous conversations, the mislabeled desk can be accurately reclassified as a cabinet. In summary, GLIS proactively employs global and local information within a chain-of-thought inference framework. 

To support high-quality global-local collaborative inference, both the local branch and the global branch are elaborately designed. For the local branch, we propose Reflected Pseudo Labels Generation (RPLG) to generate high-quality pseudo labels for training. Additionally, Background-Aware Object Localization (BAOL) is proposed to generate precise object proposals. For the global branch, the model is learned to generate scene descriptions following the demonstration of MiniGPT-v2~\cite{chen2023minigpt}.

In summary, our contributions are as follows.
\begin{itemize}
  \item We propose a lidar-based open-vocabulary detection method, GLIS, which is the first work to explore the interactions of the global scene-level information and local object-level information in this field.
  \item We introduce LLM to conduct chain-of-thought inference in the lidar-based open-vocabulary detection pipeline, where the common sense extracted from LLM is utilized to refine the detection result.
  \item To further improve the detection performance of GLIS, we propose Reflected Pseudo Labels Generation (RPLG) and Background-Aware Object Localization (BAOL), which can alleviate the influence of noises in training and testing.
\end{itemize}

Superior performance on ScanNetV2~\cite{dai2017scannet} and SUN RGB-D~\cite{song2015sun} demonstrates the effectiveness of our methods.

\section{Related Work}

\subsection{Open Vocabulary Detection}

\subsubsection{2D Open Vocabulary Detection.} 
Many works have been done to deal with 2D OVD~\cite{feng2022promptdet, gupta2022ow, yao2022detclip, du2022learning, ma2022open, zareian2021open, yao2023detclipv2, rahman2020improved, rahman2020zero, chen2023minigpt, zhong2022regionclip, zang2022open, li2022grounded, liu2023grounding, zhou2022detecting, wu2019detectron2, gao2021room}. Some works~\cite{zhou2022detecting, zareian2021open} utilize image-text pairs to conduct open-vocabulary detection. Other works~\cite{zhong2022regionclip,zang2022open} try to use pre-aligned text-image embedding space (\eg, CLIP~\cite{radford2021learning}) to improve the detection performance. For example, RegionCLIP~\cite{zhong2022regionclip} utilizes CLIP to match captions with image regions, achieving fine-grained alignment between images and texts. Recently, with the prosperity of Generative Pre-training Transformer (GPT), some works~\cite{li2022grounded, chen2023minigpt, liu2023grounding} also try to conduct 2D OVD via image grounding. GLIP~\cite{li2022grounded} proposes a scheme to unify the pre-training of object detection and image grounding. Grounding DINO~\cite{liu2023grounding} utilizes grounded pre-training to equip the closed-set detector DINO~\cite{zhang2022dino} with open-vocabulary detection abilities. Minigpt-v2~\cite{chen2023minigpt} builds a unified LLM for completing various vision-language tasks including object detection and image caption.

\subsubsection{3D Open Vocabulary Detection.}
As point cloud plays an important role in autonomous driving and robot vision, some researchers are paying attention to 3D OVD~\cite{lu2022open, lu2023open, cao2024coda, zhang2023opensight, zhang2023fm, wang2024ov}. OV-3DET~\cite{lu2022open,lu2023open} utilizes pre-trained 2D  detectors to generate 2D pseudo labels from RGB images, which are then converted to 3D pseudo labels via projection. Besides, a debiased contrastive learning strategy is proposed in OV-3DET to bridge connections between texts, images and point clouds. Unlike OV-3DET, CoDA~\cite{cao2024coda} trains a based detector with limited annotations for localization and discovers novel objects with 3D box geometry priors and 2D semantic open-vocabulary priors. FM-OV3D~\cite{zhang2023fm} obtains knowledge from pre-trained foundation models to improve the detection performance. OpenSight~\cite{zhang2023opensight} explores 3D OVD in outdoor scenes. 

However, all previous methods for lidar-based OVD only focus on the usage of object-level features, ignoring the value of the scene-level information.

\subsection{3D Object Detection}
As a basic technique for 3D world perception, lidar-based 3D object detection has achieved great progress in recent years ~\cite{qi2019deep, xie2020mlcvnet, ma2021delving, zhang2020h3dnet, xie2021venet, cheng2021back, shi2020pv, li2021lidar, qi2018frustum, shi2019pointrcnn, yang20203dssd, fan2021rangedet, lang2019pointpillars, sun2021rsn, wang2020pillar, yin2021center, zhou2018voxelnet, huang2020epnet, yang2018pixor, chen2017multi, liu2021group, misra2021end, luo20223d, fu2024eliminating}. Some works~\cite{yang2018pixor, chen2017multi} project 3D point clouds onto the Bird’s Eye View, so they can be processed by 2D CNNs. Other works~\cite{lang2019pointpillars, shi2019pointrcnn, li2021lidar, yang20203dssd} design networks for direct feature extraction from point clouds. For example,  PointRCNN~\cite{shi2019pointrcnn} proposes a two-stage detection pipeline to handle proposal generation and refining respectively. With the rise of Transformer~\cite{vaswani2017attention}, transformer-based 3D detectors~\cite{liu2021group, misra2021end} are also devised. Group-Free~\cite{liu2021group} utilizes the attention mechanism of Transformer to directly learn the contribution of each point to a single object. 3DETR~\cite{misra2021end} builds an end-to-end transformer-based detector for 3D object detection.

However, these works are focused on closed-set lidar-based 3D object detection. In contrast, our GLIS is targeted for 3D open-vocabulary object detection.

\subsection{Large Language Models}
Trained on numerous texts from the Internet in a self-supervised way, Large Language Models (LLMs) acquire the ability to generate human-like natural language~\cite{brown2020language, touvron2023llama}. For example, GPT-3~\cite{brown2020language} can chat with users or write essays according to instructions. To apply pre-trained LLMs in specific areas, many fine-tuning methods are proposed, \eg P-Tuning~\cite{liu2022p, liu2021p} and Lora~\cite{hu2021lora}. There are also works~\cite{wei2022chain, kojima2022large} trying to improve the inference ability of LLMs. For example, ~\cite{wei2022chain} utilizes the technique of Chain-of-Thought Prompting (CoT) to help LLMs conduct reasoning step by step. To introduce the advantages of LLMs from texts to other modals, Multi-Modal Large Language Models (MLLMs)~\cite{zhang2023video, liu2024visual} are proposed. For example, LLaVA~\cite{liu2024visual} equips LLMs with visual encoders for visual and language understanding.

In this paper, we devise a global-local collaborative inference scheme to utilize the inference ability of LLMs in  lidar-based OVD.

\section{Methods}

\subsection{Notation and Preliminaries}
Generally, a training sample for lidar-based OVD is consisted of three parts: the point cloud $P$, the image $I$, and the projection matrix $M$. The point cloud $P$ is a set of 3D points $\{(x_i,y_i,z_i)\}_{i=1}^{N_p}$, where $N_p$ is the points number. The 2D RGB image $I\in \mathbb{R}^{h\times w\times 3}$ is in pair with $P$. Following the lidar-based OVD experiment setup~\cite{lu2022open, lu2023open, zhang2023fm, zhang2023opensight}, images are only utilized during the training stage, while test is purely based on point clouds. The projection matrix $M$ is used for bounding box conversion between 2D and 3D. It should be noted that OVD is an unsupervised task, and no ground-truth label is provided in the training stage.

We record a bounding box as $(x,y,z,l,w,h,\theta)$, where $(x,y,z)$ is the center, $l,w,h$ are length, width, height respectively, and $\theta$ is the heading angle. We also use a 3D backbone to extract local feature $f_{loc} \in\mathbb{R}^{N_{q}\times D_p}$ and global feature $f_{glob}\in \mathbb{R}^{1\times D_p}$ from the point cloud, where $N_{q}$ is the query number of Transformer in 3D backbone, and $D_p$ is the 3D feature dimension.

\subsection{Overview}
\begin{figure}[tb]
  \centering
  \includegraphics[height=8.7cm]{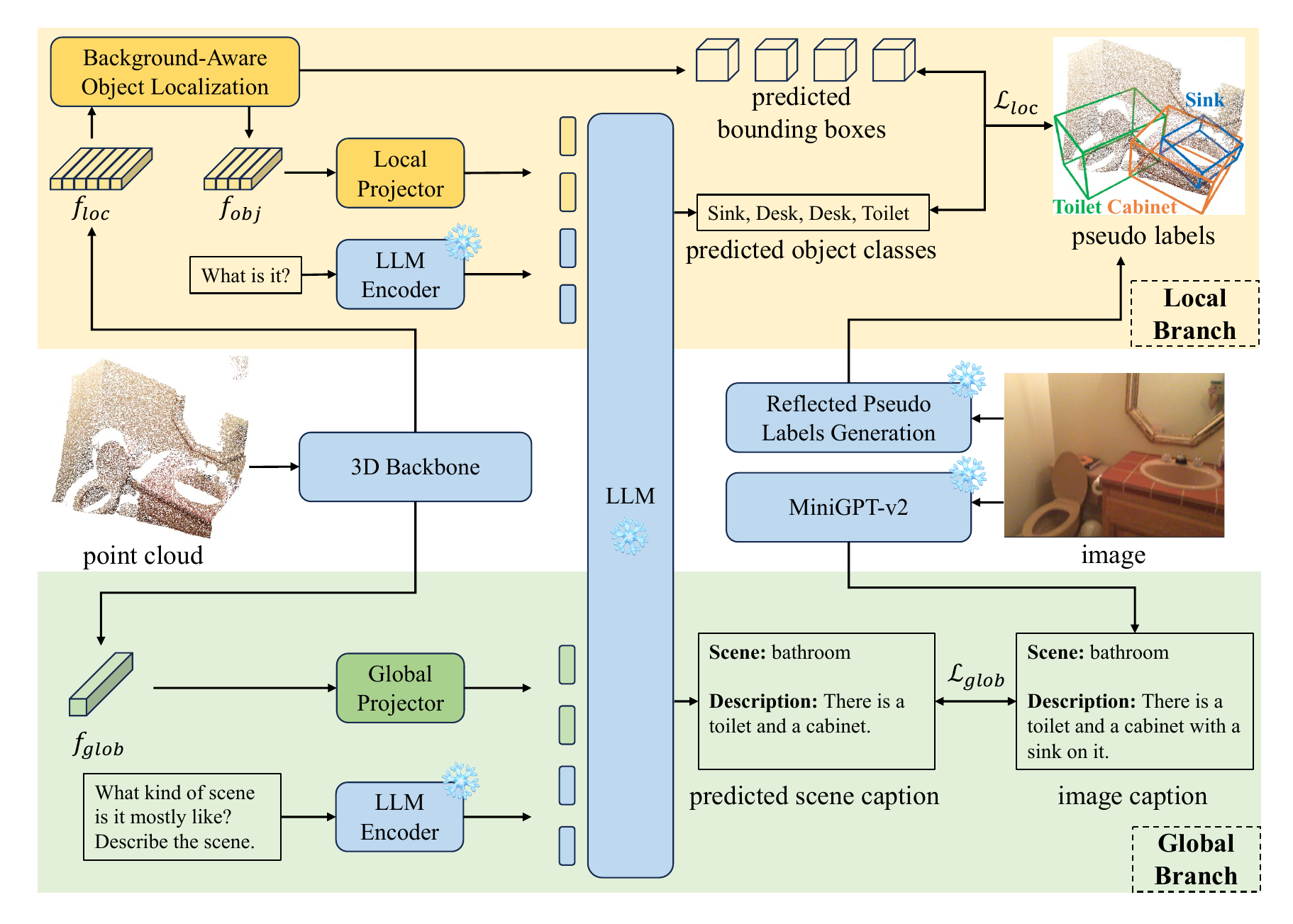}
  \caption{
  The training pipeline of GLIS.
  }
  \label{fig:overview_train}
\end{figure} 

For lidar-based open-vocabulary object detection, the object-level information and scene-level information are both useful. In this way, we propose a \textbf{G}lobal-\textbf{L}ocal Collaborative \textbf{I}nference \textbf{S}cheme (GLIS) for lidar-based open-vocabulary object detection. The whole GLIS is comprised of a training pipeline and an inference pipeline.  

\cref{fig:overview_train} shows the training pipeline of GLIS. Initially, the point cloud is input into the 3D backbone to extract local feature $f_{loc}$ and global feature $f_{glob}$. Since the unsupervised OVD task lacks supervisory signals, both local and global branches contain model training and pseudo-label acquisition. For the local branch, the bounding boxes are extracted by the Background-Aware Object Localization (BAOL) module, while the object classes are predicted by the LLM. Note that a Reflected Pseudo Labels Generation (RPLG) module is designed to generate pseudo labels, which are used as the supervised signals for loss computing. For the global branch, the scene caption is generated by the LLM based on the global feature. The scene caption is supervised by the image caption from MiniGPT-v2~\cite{chen2023minigpt}. Note that the snow symbol in \cref{fig:overview_train} means that model parameters are fixed during training. 

The inference pipeline of GLIS is presented in \cref{fig:overview_test}. Firstly, the LLM generates a description of the scene and predicts what type the scene is. Then a preliminary detection result is formed based on the local feature. Finally, chain-of-thought prompts are used to guide the LLM's inference,  where the detection result is refined accordingly. The detailed process will be introduced in section \cref{sec:inference}.

\subsection{Local Object Localization and Classification}
The local branch is responsible for the generation of preliminary detection results. Firstly, the Background-Aware Object Localization (BAOL) module extracts object proposals $\{b_i,f_{obj}^i\}_{i=1}^{N_{obj}}$ from the local feature $f_{loc}$:
\begin{equation}
\{b_i,f_{obj}^i\}_{i=1}^{N_{obj}} = \text{BAOL}(f_{loc}),
\end{equation}
where $b_i$ is the i-th object bounding box, $f_{obj}^i$ is the i-th object feature, $N_{obj}$ is the object number. Then object classes $\{c_{i}\}_{i=1}^{N_{obj}}$ are predicted by LLM:
\begin{equation}
c_i = \text{LLM}(T_{loc}, \text{Local-Projector}(f_{obj}^i)),
\end{equation}
where $c_i$ is the predicted class of i-th object and $T_{loc}$ is the LLM prompt. In practice, we use \textit{"What is it?"} as $T_{loc}$. The Local Projector is a linear layer, which aligns object features  $\{f_{obj}^i\}_{i=1}^{N_{obj}}$ to the LLM embedding space. The training of BAOL and Local Projector is supervised by the pseudo labels from the Reflected Pseudo Labels Generation (RPLG) module. In the following paragraph, we will introduce the proposed RPLG and BAOL in detail.

\subsubsection{Reflected Pseudo Labels Generation (RPLG).} As an unsupervised task, OVD lacks off-the-shelf labels for model training. To deal with this problem, previous works~\cite{lu2022open,lu2023open,zhang2023fm,zhang2023opensight} firstly obtain 2D pseudo labels from the image by 2D open-vocabulary detectors. Then these 2D pseudo labels are converted to 3D pseudo labels via the 2D-to-3D projection matrix. However, due to the limited detection ability, 2D open-vocabulary detectors may generate false labels, which may further confuse the OVD training. To alleviate this issue, we propose a reflected pseudo labels generation scheme to reduce the noise in pseudo labels. Specifically, Detectron2~\cite{wu2019detectron2} is used to generate 2D labels from images and CLIP~\cite{radford2021learning} is further adopted to play the role of reflection, \ie, checks the correctness of these labels. 

Specifically, initial 2D labels $\{\bar b_{2D}^i,\bar c_i\}_{i=1}^{\bar N}$ are generated by Detectron2, where $\bar b_{2D}^i$ is the i-th 2D bounding box, $\bar c_i$ is the i-th object class, $\bar N$ is the number of 2D labels. Then corresponding patches $\{p_i\}_{i=1}^{\bar N}$ are cropped from image $I$ according to $\{\bar b_{2D}^i\}_{i=1}^{\bar N}$. To apply CLIP for label checking, we set two prompt templates:
\begin{center}
$T^+(class)$: \textit{"This is a \{class\}.",}\\
$T^-(class)$: \textit{"This is not a \{class\}.".}
\end{center}
These templates, together with patches $\{p_i\}_{i=1}^{\bar N}$ and classes $\{\bar c_i\}_{i=1}^{\bar N}$, are sent into CLIP to calculate confidence scores:
\begin{equation}
[\phi_i^+,\phi_i^-] = \text{Softmax}(\text{CLIP}(T^+(\bar c_i),p_i),\text{CLIP}(T^-(\bar c_i),p_i)),
\end{equation}
where $\phi_i^+$ is the confidence score that $p_i$ belongs to class $c_i$ and vice versa. We keep labels with $\phi_i^+$ higher than a predefined threshold $\phi_{CLIP}$, creating new 2D labels $\{\tilde{b}_{2D}^i,\tilde{c}_i\}_{i=1}^{\tilde N}$. These 2D labels are converted to 3D pseudo labels $\{\tilde{b}_i,\tilde{c}_i\}_{i=1}^{\tilde N}$ via projection matrix $M$, where $\tilde N$ is the number of 3D pseudo labels.

\subsubsection{Background-Aware Object Localization (BAOL).} 
Due to noises within the point cloud, the detector may confuse foreground objects with the background and outputs false object proposals. To select high-quality object proposals, we propose background-aware object localization . Basically, class-agnostic object proposals $\{\hat {b}_i,\hat o_i\}_{i=1}^{N_q}$ are extracted from $f_{loc}$ via a series of prediction heads, where $\hat b_i$ is the i-th bounding box, and $\hat o_i$ is the confidence of the i-th proposal. Then proposals whose confidence is below a threshold $\phi_{obj}$ are removed, forming the class-agnostic detection result $\{{b}_i,o_i\}_{i=1}^{N_{obj}}$, where $N_{obj}$ is the number of objects. 

In closed-vocabulary detection, the prediction of confidence can be learned from groundtrurh labels. However, in OVD, we have to design the supervision signals manually. Preliminarily, we conduct bipartite matching between the proposals $\{\hat b_i\}_{i=1}^{N_q}$ and the pseudo labels $\{\tilde b_i\}_{i=1}^{\tilde N}$ based on IoU. Then all proposals are labeled via:
\begin{equation}
y_i = \begin{cases}
1 & \exists j,\hat{b}_i \text{ is matched with } \tilde{b}_j,\\
0 & \text{otherwise},
\end{cases}
\end{equation}
where $y_i=1$ means the i-th proposal is positive, \ie, is a foreground object, and vice versa. 
However, such label assignment can be inaccurate in two situations: (i) The matched proposal has a low IoU with the matched label, which means the proposal is not accurate enough; (ii) Two different proposals refer to the same object, yet only one of the proposal could be labeled as foreground object, which may cause confusion. To tackle these situations, we further modify the label rule. For situation (i), if a proposal has a IoU below $\phi_{low}$ with the matched pseudo bounding box, then it will be labeled as negative. For situation (ii), if the IoU of the two proposals are higher than $\phi_{high}$, then both of them will be labeled as positive. For simplicity, we still use $\{y_i\}_{i=1}^{N_q}$ to denote the refined labels. 

\subsection{Global Scene Understanding and Description}
The global branch predicts scene type $s$ (\eg, bedroom, kitchen, \etc) and generates scene description $d$ based on global feature $f_{glob}$. Specifically, we prompt LLM with the following text:
\begin{center}
$T_{glob}$: \textit{"What kind of scene is it mostly like? Describe the scene.".}
\end{center}
Besides, we use the global projector to align global feature $f_{glob}$ to the LLM embedding space. The whole process could be represented as 
\begin{equation}
s,d = \text{LLM}(T_{glob}, \text{Global-Projector}(f_{glob})).
\end{equation}

For supervision, we utilize the 2D-based MiniGPT-v2~\cite{chen2023minigpt} to generate scene type label $\tilde s$ and scene description label $\tilde d$ from the paired image, which are used for loss computation with the LLM answer.
\subsection{Global-Local Collaborative Inference with LLM}
\label{sec:inference}
\begin{figure}[tb]
  \centering
  \includegraphics[height=8.3cm]{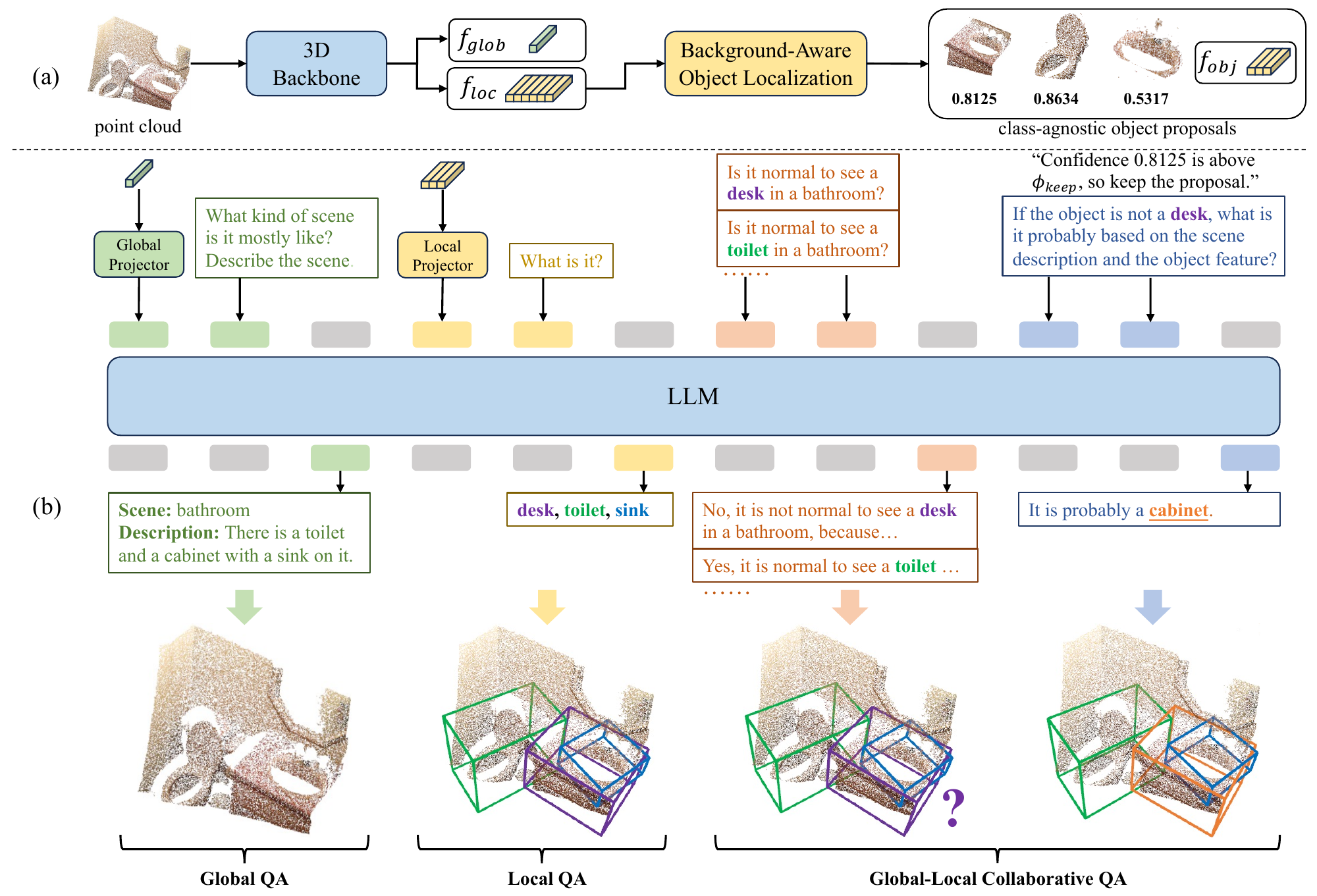}
  \caption{The inference pipeline of GLIS.
  }
  \label{fig:overview_test}
\end{figure} 

In this section, we will introduce the inference pipeline of our proposed GLIS, which is presented in \cref{fig:overview_test}. Firstly, as shown in \cref{fig:overview_test}\textcolor{red}{a}, global feature $f_{glob}$ and local feature $f_{loc}$ are extracted from 3D backbone. Then class-agnostic object proposals, as well as their confidence scores and 3D features $f_{obj}$, are generated by the BAOL module.

With the extracted local and global information, we can conduct global-local collaborative inference (GLCI) with LLM and refine the detection result accordingly. Generally, as shown in \cref{fig:overview_test}\textcolor{red}{b}, chain-of-thought prompts are used to guide the LLM's inference. The whole inference process could be divided into three stages, \ie, global QA, local QA, and global-local collaborative QA. In global QA, the LLM is asked to predict the scene type and describe the scene based on the global feature $f_{glob}$. In local QA, $f_{obj}$ is utilized to predicted the classes of the class-agnostic object proposals. 

After finishing global QA and local QA, a preliminary detection result is given. This detection result will be refined step by step in global-local collaborative QA. Specifically, for each predicted class, the following prompt template is used to check its rationality:
\textit{"Is it normal to see a \{class\} in a \{scene type\} ?".}
If a class $c$ is thought as reasonable to exist in the predicted scene type, then all predicted objects belonging to class $c$ are reserved, \eg, toilet in \cref{fig:overview_test}\textcolor{red}{b}. Contrastly, those objects that are not likely to be in the scene will be further inspected. Specifically, for each unreasonable object, if its confidence is below a predefined threshold $\phi_{keep}=0.75$, it will be automatically removed, and vice versa. For example, in \cref{fig:overview_test}\textcolor{red}{b}, the object proposal of desk is kept as its confidence is beyond $\phi_{keep}$. To obtain the correct class of this proposal, we further prompt LLM with following template:
\textit{"If the object is not a desk, what is it probably based on the scene description and the object feature?".}
Then the LLM utilizes the information from both local and global to give the answer. Finally, the object class is corrected according to the LLM's answer.

\subsection{Training Objectives}
The training loss contains four parts, \ie, bounding box regression loss $\mathcal{L}_{bbox}$, confidence prediction loss $\mathcal{L}_{conf}$, object classifiction loss $\mathcal{L}_{cls}$, and scene understanding loss $\mathcal{L}_{scene}$. We use the regression loss function in 3DETR~\cite{misra2021end} to compute $\mathcal{L}_{bbox}$, which could be represented as:
\begin{equation}
\mathcal{L}_{bbox} = \text{RegressionLoss}(\{b_i\}_{i=1}^{N_{obj}}, \{\tilde{b}_i\}_{i=1}^{\tilde N}).
\end{equation}
The confidence prediction loss is defined as:
\begin{equation}
\mathcal{L}_{conf} = -\frac{1}{N_q}\sum_{i=1}^{N_{q}}[y_i\log \hat{o}_i+\lambda_{conf}(1-y_i)\log(1-\hat{o}_i)],
\end{equation}
where $\lambda_{conf}$ is a balanced factor. As object classes are predicted by LLM, we compute $\mathcal{L}_{cls}$ by maximizing the possibility of label text tokens~\cite{xu2023pointllm}. Specifically, assuming the label text is a sequence of tokens $t=(w_1,w_2,\cdots,w_l)$ and the predicted possibility for each token is $p(t)=[p(w_1),p(w_2),\cdots,p(w_l)]$, the text loss is defined as:
\begin{equation}
\mathcal{L}_{text}(p(t))=-\sum_{i=1}^l \log p(w_i).
\end{equation}
In this way, $\mathcal{L}_{cls}$ could be computed as:
\begin{equation}
    \mathcal{L}_{cls} = \frac{1}{\tilde N}\sum_{i=1}^{\tilde N}\mathcal{L}_{text}(p_{loc}(\tilde {c}_i)),
\end{equation}
where $p_{loc}$ is the LLM predicted token possibility in the local branch. Similarly, $\mathcal{L}_{scene}$ could be computed via:
\begin{equation}
\mathcal{L}_{scene} = \mathcal{L}_{text}(p_{glob}(\tilde s)) + \mathcal{L}_{text}(p_{glob}(\tilde {d})),
\end{equation}
where $p_{glob}$ is the LLM predicted token possibility in the global branch. 

In summary, the total loss is computed as
\begin{equation}
\mathcal{L}=\lambda_1\mathcal{L}_{bbox} + \lambda_2\mathcal{L}_{conf} + \lambda_3\mathcal{L}_{cls}+\lambda_4\mathcal{L}_{scene},
\end{equation}
where $\lambda_1,\lambda_2,\lambda_3, \lambda_4$ are balanced factors.

\section{Experiments}

\subsection{Datasets and Metrics}
We conduct experiments on two datasets: ScanNetV2~\cite{dai2017scannet} and SUN RGB-D~\cite{song2015sun}.

\subsubsection{ScanNetV2} is a widely used 3D object detection and semantic segmentation dataset, which has 1513 scenes with over 200 object classes.
\subsubsection{SUN RGB-D} is a large 3D object detection and scene understanding dataset, which contains 10335 samples with around 800 object classes. 
\subsubsection{Metrics.} We use the mean Average Precision (mAP) at the IoU threshold of 0.25 to evaluate the detection performance. For a fair comparison, we evaluate our GLIS on the top-20 object classes in ScanNetV2 and SUN RGB-D respectively, following OV-3DET~\cite{lu2023open}. We also report the results on top-10 classes for comparison with methods like ~\cite{lu2022open,zhang2023fm}. The metrics are notated as $mAP_{25}^{20cls}$ and $mAP_{25}^{10cls}$ respectively for distinguishing.
\subsection{Implementation Details}
The training process contains two phases: 1) training the 3D backbone and the bounding box prediction heads; 2) training the object confidence prediction head, as well as the local projector and global projector. The training of phase 1 lasts for 400 epochs with a total batch size of 32 (\ie, a single batch size 4 $\times$ 8 GPUs). The training of phase 2 lasts for 50 epochs with a total batch size of 16. The base learning rate is set as $1e-4$. The loss balanced factors are $\lambda_{conf}=0.2,\lambda_1=4,\lambda_2=10,\lambda_3=1,\lambda_4=1$, respectively. The mentioned thresholds are $\phi_{CLIP}=0.5,\phi_{obj}=0.1,\phi_{low}=0.25,\phi_{high}=0.6,\phi_{keep}=0.75$, respectively. We choose the model architecture of 3DETR~\cite{misra2021end} as our 3D backbone and bounding box prediction heads. We use LLaMA~\cite{touvron2023llama} as
the LLM backbone, which is initialized by the checkpoint vicuna-7b-v1.5-16k~\cite{chiang2023vicuna}. All experiments are conducted on 8 A800 GPUs. 
\subsection{Main Results}
\begin{table}[tb]
  \caption{Comparisons with other methods on ScanNetV2}
  \label{tab:scannet}
  \centering
  \renewcommand\arraystretch{0.5}
  \resizebox{\textwidth}{15mm}{
  \begin{tabular}{c|c|ccccccccccccccccccccc}
    \toprule
    Method&$mAP_{25}^{10cls}$&toilet&bed&chair&sofa&dresser&table&cabinet&bookshelf&pillow&sink\\
    \midrule
    OV-3DETIC~\cite{lu2022open} & 12.65 &48.99&2.63&7.27&18.64&2.77&14.34&2.35&4.54&3.93&21.08\\
    FM-OV3D~\cite{zhang2023fm}&21.53&62.32&41.97&22.24&31.80&1.89&10.73&1.38&0.11&12.26&30.62\\
    OV-3DET~\cite{lu2023open}&24.36&57.29&42.26&27.06&31.50&8.21&14.17&2.98&5.56&23.00&31.60\\
    CoDA~\cite{cao2024coda}&28.76&68.09&44.04&28.72&44.57&3.41&20.23&5.32&0.03&27.95&45.26\\
    GLIS (ours)&\bf{30.94}&73.90&39.69&39.51&44.41&6.09&25.38&5.92&8.31&25.63&43.51\\
    \midrule
Method&$mAP_{25}^{20cls}$&bathtub&refrigerator&desk&nightstand&counter&door&curtain&box&lamp&bag\\
    \midrule
    OV-3DET~\cite{lu2023open}&18.02&56.28&10.99&19.72&0.77&0.31&9.59&10.53&3.78&2.11&2.71&\\
    CoDA~\cite{cao2024coda}&19.32&50.51&6.55&12.42&15.15&0.68&7.95&0.01&2.94&0.51&2.02\\
    GLIS (ours)&\bf{20.83}&53.21&4.76&20.79&7.62&0.09&0.95&7.79&3.32&3.73&1.93\\   
  \bottomrule
  \end{tabular}}
\end{table}

\begin{table}[tb]
  \caption{Comparisons with other methods on SUN RGB-D}
  \label{tab:sunrgbd}
  \centering
\resizebox{\textwidth}{15mm}{
  \begin{tabular}{c|c|ccccccccccccccccccccc}
    \toprule
Method&$mAP_{25}^{10cls}$&~toilet~&~~bed~~&chair&~bathtub~&~~~sofa~~~&dresser&scanner&fridge&~lamp~&desk\\
\midrule
OV-3DETIC~\cite{lu2022open}&13.03&43.97&6.17&0.89&45.75&2.26&8.22&0.02&8.32&0.07&14.60\\
FM-OV3D~\cite{zhang2023fm}&21.47&55.00&38.80&19.20&41.91&23.82&3.52&0.36&5.95&17.40&8.77\\
OV-3DET~\cite{lu2023open}&\bf{31.06}&72.64&66.13&34.80&44.74&42.10&11.52&0.29&12.57&14.64&11.21\\
GLIS (ours)&30.83&69.88&63.83&34.78&49.62&40.78&10.73&1.49&8.37&16.40&12.44&\\
\midrule
Method&$mAP_{25}^{20cls}$&~table~ &~~stand~~&cabinet&~counter~&~~~bin~~~&bookshelf&pillow&microwave&~sink~&stool\\
\midrule
OV-3DET~\cite{lu2023open}&20.46&23.31&2.75&3.40&0.75&23.52&9.83&10.27&1.98&18.57&4.10\\
GLIS (ours)&\bf{21.45}&19.17&13.84&2.75&0.59&22.22&12.65&15.78&5.30&27.62&0.84\\

  \bottomrule
  \end{tabular}}
\end{table}

\subsubsection{ScanNetV2.} As shown in \cref{tab:scannet}, our proposed GLIS greatly improves the open-vocabulary detection performance on ScanNetV2. Compared to previous sota method CoDA~\cite{cao2024coda}, $mAP_{25}^{10cls}$ is raised from $28.76\%$ to $30.94\%$ and $mAP_{25}^{20cls}$ is raised from $19.32\%$ to $20.83\%$. Our methods also significantly improve the detection precision of many classes, \eg, chair is improved by $10.79\%$, toilet is improved by $5.81\%$, and table is improved by $5.15\%$. 
\subsubsection{SUN RGB-D.} The detection precision on SUN RGB-D is reported in \cref{tab:sunrgbd}. When tested on 20 classes, $mAP_{25}^{20cls}$ is improved from $20.46\%$ to $21.45\%$, demonstrating the effectiveness of our methods. Specifically, stand is improved by $11.09\%$, sink is improved by $9.05\%$, and bathtub is improved by $4.88\%$, \etc. Our methods also achieve competitive performance compared to OV-3DET~\cite{lu2023open} when evaluated on 10 classes. 

\subsection{Ablation Study}
\begin{table}[tb]
  \caption{Ablation study on ScanNetV2}
  \label{tab:ablation}
  \centering

  \begin{tabular}{ccccc}
    \toprule
Method&$mAP_{25}^{10cls}$&~~~~~~~~~~~~~~~~~~&$mAP_{25}^{20cls}$&~~~~~~~~~~~~~~~~~~\\
\midrule
Base Model &28.59&&19.36\\
+RPLG&28.80&\textcolor[RGB]{0,128,0}{+0.21}&19.51&\textcolor[RGB]{0,128,0}{+0.15}\\
+BAOL&29.75&\textcolor[RGB]{0,128,0}{+0.95}&20.03&\textcolor[RGB]{0,128,0}{+0.52}\\
+GLCI&\bf{30.94}&\textcolor[RGB]{0,128,0}{+1.19}&\bf{20.83}&\textcolor[RGB]{0,128,0}{+0.80}\\

  \bottomrule
  \end{tabular}
\end{table}

To analyze the effect of each module in our proposed GLIS, we conduct an ablation study on ScanNetV2. Detailed results are presented in \cref{tab:ablation}

\subsubsection{Base Model.} Base model is different from GLIS in three aspects: i) In RPLG, all pseudo labels from Detectron2 are reserved. ii) In BAOL, all local features are reserved (\ie, $f_{obj}=f_{loc}$). iii) Global inference is not performed.
\subsubsection{Effect of RPLG.} With RPLG added to the Base Model, the detection precision increases. For example, $mAP_{25}^{10cls}$ is improved from $28.59\%$ to $28.80\%$. Such improvement demonstrates that RPLG can recognize false detection in MiniGPT-v2 and improve the pseudo labels' quality.  
\subsubsection{Effect of BAOL.} BAOL further refines the detection performance. For instance, $mAP_{25}^{10cls}$ is lifted from $28.80\%$ to $29.75\%$ and $mAP_{25}^{20cls}$ is lifted from $19.51\%$ to $20.03\%$. This proves that BAOL can overcome the disturbance of noises in point clouds, resulting in better localization for interested objects.
\subsubsection{Effect of GLCI.} As shown in \cref{tab:ablation}, GLCI significantly improves the detection performance. Specifically, $mAP_{25}^{10cls}$ is increased by $1.19\%$ and $mAP_{25}^{20cls}$ is increased by $0.80\%$. Such results show that GLIS successfully utilizes the information from both local and global, and refines the detection result with effective inference.  
\subsection{Visualizations}
\label{vis}
\begin{figure}[tb]
  \centering
  \includegraphics[height=7.6cm]{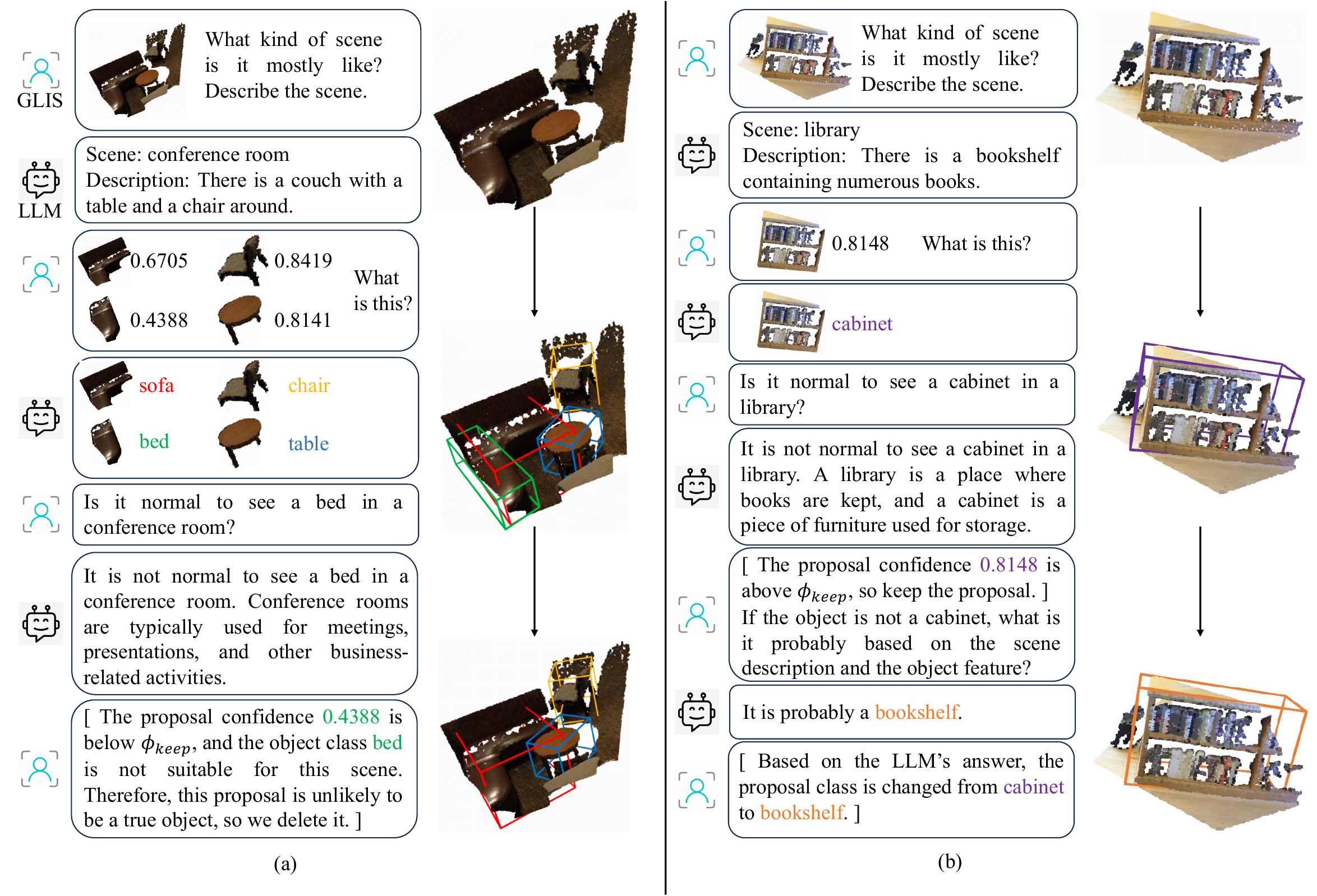}
  \caption{Visualizations of GLIS. The score of each proposal is the confidence that the proposal is truly a foreground object. These proposals, as well as their confidence scores, are generated by BAOL.
  }
  \label{fig:vis}
\end{figure} 

To further show the effectiveness of our proposed GLIS, we present visualization results in \cref{fig:vis}. The visualization mainly exhibits the dialogue between GLIS and the LLM. Besides, the detection result of each stage is presented, which changes according to the dialogue. 

As shown in \cref{fig:vis}\textcolor{red}{a}, LLM recognizes the scene as a conference room based on global information. Then, four class-agnostic object proposals are generated by BAOL. The four proposals are recognized as a sofa, a chair, a bed, and a table respectively by LLM. Note that the score (\eg, 0.6705) of each proposal is the confidence that the proposal is truly a foreground object. So far, an initial detection result has formed. Subsequently, the global-local collaborative inference is conducted with predefined questions. For example, the question \textit{"Is it normal to see a bed in a conference room?"} is used to guide the LLM's inference. Then LLM outputs the answer to the question, \ie, \textit{"It is not normal to see a bed in a conference room."}, and gives reasonable explanations. Based on the LLM's answer, our scheme automatically recognizes the bed as false detection. Subsequently, as the confidence of the bed is below the predefined threshold $\phi_{keep}=0.75$, the wrongly detected bed is automatically deleted from the result. Such example shows that GLIS can effectively conduct inference with global and local information, extracting common sense from LLM to eliminate wrongly detected objects.   

\cref{fig:vis}\textcolor{red}{b} exhibits another example. Initially, LLM recognizes the scene as a library, and generates scene description \textit{"There is a bookshelf containing numerous books."}. Only one object proposal is generated from BAOL, and the proposal is recognized as a cabinet. Then, our scheme automatically asks the LLM with the question \textit{"Is it normal to see a cabinet in a library?"}, and LLM answers that \textit{"It is not normal to see a cabinet in a library."}. As the proposal confidence $0.8148$ is above the threshold $\phi_{keep}=0.75$, the proposal is not removed. To correct the predicted class of this proposal, our scheme asks LLM to change the object class according to the scene description and the object feature. Based on local and global information, LLM corrects the object class from cabinet to bookshelf. 

\begin{figure}[tb]
  \centering
  \includegraphics[height=3.2cm]{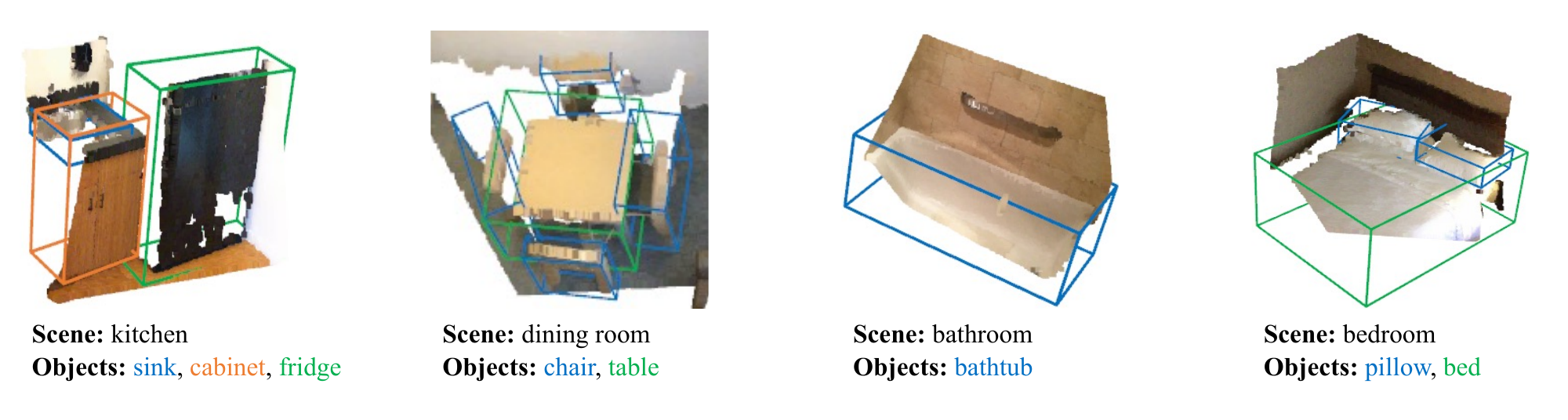}
  \caption{Visualizations of detection results.
  }
  \label{fig:result}
\end{figure}

More visualizations of detection results are presented in \cref{fig:result}.

\section{Conclusion and Discussion}
In this paper, we propose GLIS for lidar-based open-vocabulary detection task, which is the first to introduce global-local collaboration in this area. The proposed GLIS extracts object-level information from the local branch and scene-level information from the global branch. With local and global information, a pre-trained LLM is utilized for inference to refine the detection result. To further improve the performance, RPLG and BAOL are devised to ameliorate the object-level information. Experiments on ScanNetV2 and SUN RGB-D demonstrate the effectiveness of GLIS, which achieves state-of-the-art results.

\subsubsection{Limitation.} The limitation of GLIS exists due to the noises within the point cloud and the false pseudo labels generated from the 2D image. Though we have proposed methods to alleviate the influence, these noises may still confuse the model. These limitations could inspire our future work.

\section*{Acknowledgements} This research is supported in part by National Science and Technology Major Project (2022ZD0115502), National Natural Science Foundation of China (NO. 62122010, U23B2010), Zhejiang Provincial Natural Science Foundation of China (Grant No. LDT23F02022F02), Beijing Natural Science Foundation (NO. L231011), and Beihang World TOP University Cooperation Program.

\bibliographystyle{splncs04}
\bibliography{egbib}
\end{document}